\documentclass{article}
\usepackage{spconf,amsmath,graphicx}

\usepackage{enumitem}
\setlist{nosep, leftmargin=14pt}

\usepackage{mwe} % to get dummy images

\title{Utilizing mask-guided cross-image attention for zero-shot \textit{in-silico} histopathologic image generation with a diffusion model}
\name{
  \small
  \begin{tabular}{@{}c@{}}
  Dominik Winter$^{\star}$ 
  \qquad
  Nicolas Triltsch$^{\star}$ 
  \qquad
  Marco Rosati$^{\star}$ 
  \qquad
  Anatoliy Shumilov$^{\star}$ 
  \qquad
  Ziya Kokaragac$^{\star}$ 
  \qquad
  Yuri Popov$^{\star}$
  \\
  Thomas Padel$^{\star}$ 
  \qquad
  Laura Sebastian Monasor$^{\star}$ 
  \qquad
  Ross Hill$^{\dagger}$ 
  \qquad
  Markus Schick$^{\star}$ 
  \qquad
  Nicolas Brieu$^{\star}$
  \end{tabular}
}

\address{\small$^{\star}$ AstraZeneca Computational Pathology GmbH, Bernhard-Wicki-Str. 5, 80636 Munich, Germany \\
\small$^{\dagger}$AstraZeneca, Discovery Centre, 1 Francis Crick Avenue, Cambridge, CB2 0AA, United Kingdom}
\begin{document}
%\ninept
%%
\maketitle

\begin{abstract}
Utilizing generative models to create \textit{in-silico} data presents an economical alternative to the traditional methods of staining, imaging, and annotating images in the computational pathology workflow. Specifically, appearance transfer diffusion models enable the generation of images without the need for model training, making the process swift and efficient. While originally developed for natural images, these models can transfer foreground objects from a source to a target domain, with less emphasis on the background. In computational pathology, however, every aspect of an image, including the background, can be crucial for understanding the tumor micro-environment.  
In this study, we adapted an appearance transfer diffusion model to align with the demands of computational pathology by adjusting the AdaIN feature statistics in the denoising process. The effectiveness of this modified method was demonstrated through its application to epithelium segmentation. The results demonstrated superior performance compared to the baseline approach, indicating that the number of manual annotations necessary for model training could be reduced by 75\% without sacrificing accuracy. The authors expect that this research will promote the use of zero-shot diffusion models in computational pathology.

\end{abstract}
\section{Introduction}

\begin{figure}[!htb]
  \begin{minipage}[b]{1.0\linewidth}
    \centering
    \centerline{\includegraphics[width=8.5cm]{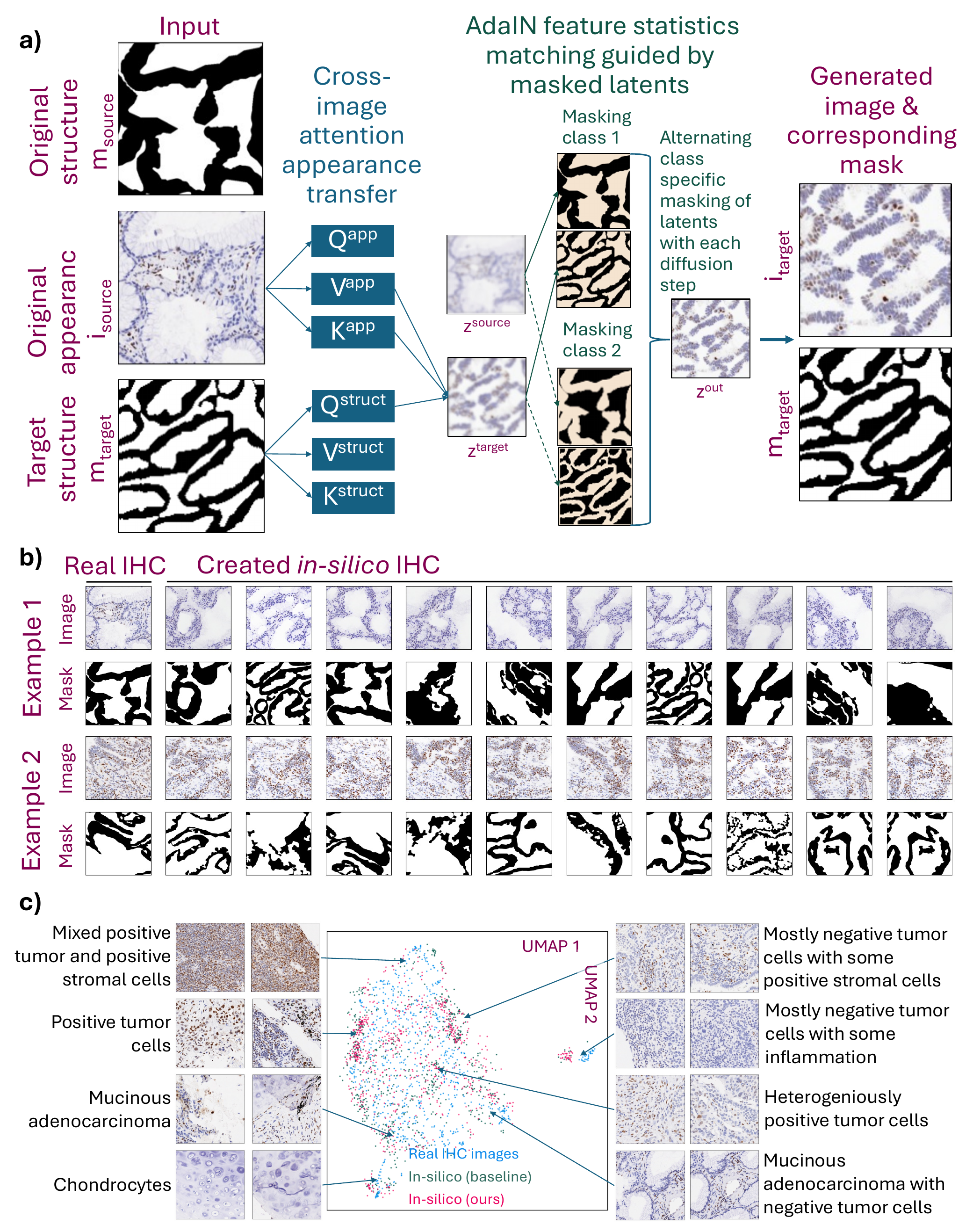}}
    \vspace*{-2mm}
    \caption{
    \textbf{a} By swapping key, values and queries from the appearance image (${i_{source}}$) with the target structure image (${m_{target}}$) we transfer the appearance of an image to the structure of a mask, generating an image (${i_{target}}$) that matches the target mask (${m_{target}}$). This process alternates AdaIN feature statistics matching across different classes, utilizing the existing segmentation masks on the latent representations ${z^{source}}$ and ${z^{target}}$. \textbf{b} Real images and their corresponding \textit{in-silico} images, which share the same appearance but different structures. \textbf{c} A UMAP plot reveals overlapping clusters of real and \textit{in-silico} images.
    }
    \vspace*{-2mm}
  \label{figure1}
\end{minipage}
\end{figure}
  
\label{sec:intro}
Developing deep learning models in computational pathology requires datasets with accurate, relevant information for tasks like segmentation and object detection. Creating these datasets involves stages such as sample preparation, staining, imaging, and expert manual annotation, which is labor-intensive and costly \cite{van2021deep}. In contrast, \textit{in-silico} data provides a more efficient, cost-effective alternative by generating datasets with the necessary properties through controllable parameterization \cite{kazerouni2023diffusion}.
Image generation significantly impacts computational pathology by enabling the creation of realistic datasets, reducing the need for costly manual annotations \cite{kazerouni2023diffusion, brieu2024auxiliary, winter2024restaingan}. Diffusion models in medicine, biology, and pathology \cite{kazerouni2023diffusion} have diverse applications, including segmentation \cite{wolleb2022diffusion}, cell shape prediction \cite{waibel2023diffusion}, and artifact recovery \cite{moghadam2023morphology}. These models, trained to remove noise from images, can generate \textit{in-silico} images from Gaussian noise. They have been used to create realistic H\&E stained images \cite{moghadam2023morphology, xu2023vit} and datasets for rare cancers and nuclei classification \cite{oh2023diffmix, shrivastava2023nasdm, kataria2024staindiffuser, yu2023diffusion}. Text-conditioned diffusion models can generate pathology images from reports \cite{yellapragada2024pathldm} or at gigapixel scales \cite{harb2024diffusion, aversa2024diffinfinite}. Our work focuses on analyzing the greater complexity of immunohistochemistry (IHC) stained images. 
The recent work of Alaluf and al. shows that the appearance between images can be transferred during the desnoising diffusion process using cross- and self-attention mechanisms \cite{alaluf2023cross}. The keys, values and queries are first projected through the learned linear projections in each self-attention layer of the network. The attention scores, which indicate the relevance of each key to the corresponding queries, are then calculated for all keys and weighted by a Softmax operation.  The aggregation of the scores results in an attention map which captures correspondences across the entire image. The queries capture the semantic information of each spatial location. The keys offer context for the queries. The values contain the content that is transferred from the source image to the target structure. 
In the case of appearance transfer for natural images, the foreground object (e.g. an animal) is of high importance in opposition to the background (e.g. the sky). However, in histopathology, both the foreground (e.g. epithelium) and the background (e.g. stroma) regions are important. Achieving a high level of realism for both is decisive to solve downstream tasks such as segmentation. 
In this study, we introduce a novel appearance transfer diffusion models for generating \textit{in-silico} data of IHC images in a zero-shot approach. By transferring the appearance of an existing image to an unpaired mask with realistic morphological structures, we generate images with high fidelity. Specifically, we aim to transfer the appearance of epithelium and stroma into different spatial organizations, while preserving the unique morphological characteristics of each tissue type. This method has the potential to generate highly realistic and diverse datasets for training and evaluating computational models in pathology.

\begin{figure}[!htb]
  \begin{minipage}[b]{1.0\linewidth}
    \centering
    \centerline{\includegraphics[width=8.5cm]{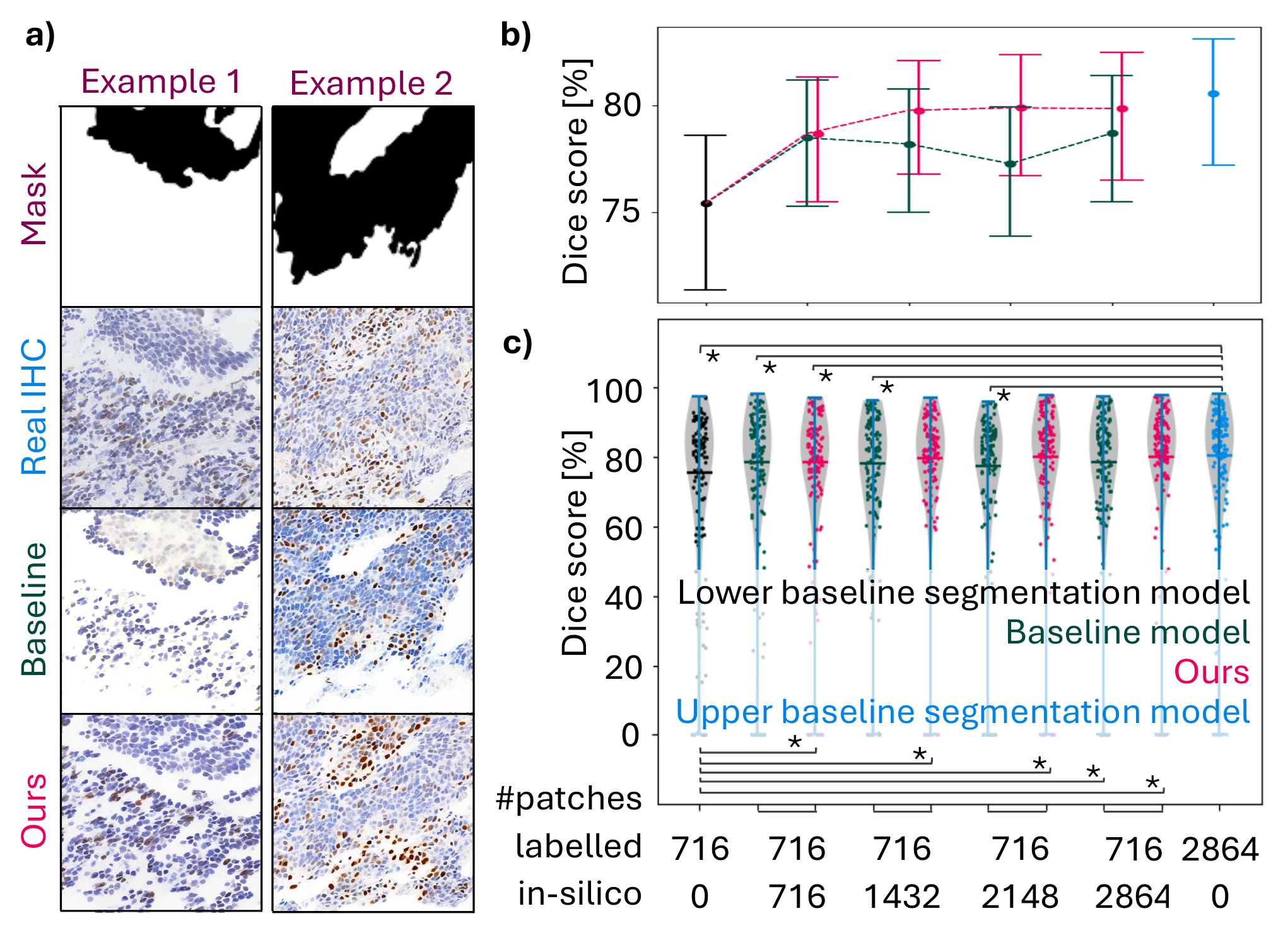}}
    \vspace*{-2mm}
    \caption{\textbf{a} Our model generates more realistic images than the baseline model. \textbf{b}  From a dataset of 2,864 manually annotated images, we use a 25\% subset (N=716 images) and generate four \textit{in-silico} images from each image.  Semantic epithelium segmentation models trained on images generated with our diffusion model (red) outperform those trained on images generated by the baseline model (petrol). The upper (blue) and lower (black) baseline segmentation models were trained on the full dataset and the 25\% subset of the labeled patches, respectively. Our test-set was annotated by three independent pathologists. The mean Dice score and corresponding confidence intervals (whiskers) are displayed. \textbf{c} A violin plot showing Dice scores for each of the 43 FOVs and the three pathologists reveals significant differences (p $<$ 0.05, Wilcoxon signed-rank test, N=129), with statistical significance indicated by a star.
    }
    \vspace*{-2mm}
\label{figure2}
\end{minipage}
\end{figure}

\section{Methods}
We adapted the baseline model developed by Alaluf et al. \cite{alaluf2023cross} to enable appearance transfer from existing IHC images to masks. This adaptation involved generating \textit{in-silico} images with realistic IHC appearance and corresponding pixel-accurate segmentation masks, derived from real IHC images. The augmented \textit{in-silico} dataset was then used to train a downstream semantic segmentation model. The paper is organized into two main sections. he first section revisits the single-class guided diffusion model originally presented by Alaluf et al. \cite{alaluf2023cross}. The second section provides a detailed description of our technical contributions, with a particular focus on the modifications made to the original approach [16] to support a multi-class setup.

\subsection{Single-Class \textit{in-silico} Image Generation}
The guided diffusion model \cite{alaluf2023cross} requires a pair of inputs: an image ${i_{source}}$ and its corresponding segmentation mask ${m_{source}}$ from the source domain, along with a segmentation mask ${m_{target}}$ from the target domain. The model then generates an \textit{in-silico} image ${i_{target}}$ in the target domain that spatially corresponds to the target segmentation mask ${m_{target}}$ (Fig. \ref{figure1}). 
To achieve this, Stable Diffusion \cite{rombach2022high} a pretrained text-to-image model, is utilized for zero-shot appearance transfer (Fig. \ref{figure1}). By replacing the keys and values of the image ${i_{source}}$ while preserving the queries from the mask ${m_{target}}$, the model enables the transfer of the appearance contained in ${i_{source}}$ to the structure of ${m_{target}}$ between semantically similar objects. 
The appearance transfer process begins with the calculation of the inverted latents $z_T^{i_{source}}$ and $z_T^{m_{target}}$. These latents are then passed through the denoising U-Net at each forward step. Notably, the standard self-attention in the U-Net decoder is replaced with cross-image attention. During each denoising step t, two forward passes are performed. The first uses the cross-attention layer:
$\epsilon^{cross}=\epsilon_\theta^{cross}(z_t^{out})$.
The second step uses the self-attention layer:
$\epsilon^{self}=\epsilon^{self} (z_t^{out})$.
The two noise predictions are then combined as follows:
$\epsilon^{t}=\epsilon^{self}+\alpha(\epsilon^{cross}-\epsilon^{self})$.
Where ${\alpha}$ represents the guidance scale, $z_{t-1}^{out}$ is the latent code at step $t-1$ and ${\epsilon}^t$ is the modified noise.
Additionally, the Adaptive Instance Normalization (AdaIN) operation is used to align the mean and variance of content features with those of style features \cite{huang2017arbitrary}, thereby matching the feature statistics between the latent representations.

\subsection{Multi-Class \textit{in-silico} Image Generation}
We have introduced modifications to the appearance transfer mechanism. First, we incorporated class-specific feature statistics matching within the AdaIN operation, guided by the existing segmentation masks ${m_{c,source}}$ and ${m_{c,target}}$ through class-specific masked latents, where c denotes the class. To achieve this, we applied a mask to the latents:
$z_{t,c}^{i_{source}}=mask(z_t^{i_{source}})$ and 
$z_{t,c}^{m_{target}}=mask(z_t^{m_{target}})$ 
followed by the AdaIN operation:
$z_{t,c}^{out}=AdaIN(z_{t,c}^{m_{target}},z_{t,c}^{i_{source}})$ 
which aligns the color distributions for each class between the generated and input images.
Additionally, we mapped the class labels to the timesteps, alternating between class-specific AdaIN masking at each forward step of the denoising process. For instance, in a two-class appearance transfer scenario, the foreground class mask was applied during every even denoising step, while the background class mask was applied during every odd denoising step. These two novel modifications significantly enhanced the fidelity and realism of the background region in the generated in-silico images, as compared to the original method \cite{alaluf2023cross}.
\begin{figure}[htb]
  \begin{minipage}[b]{1.0\linewidth}
    \centering
    \centerline{\includegraphics[width=8.5cm]{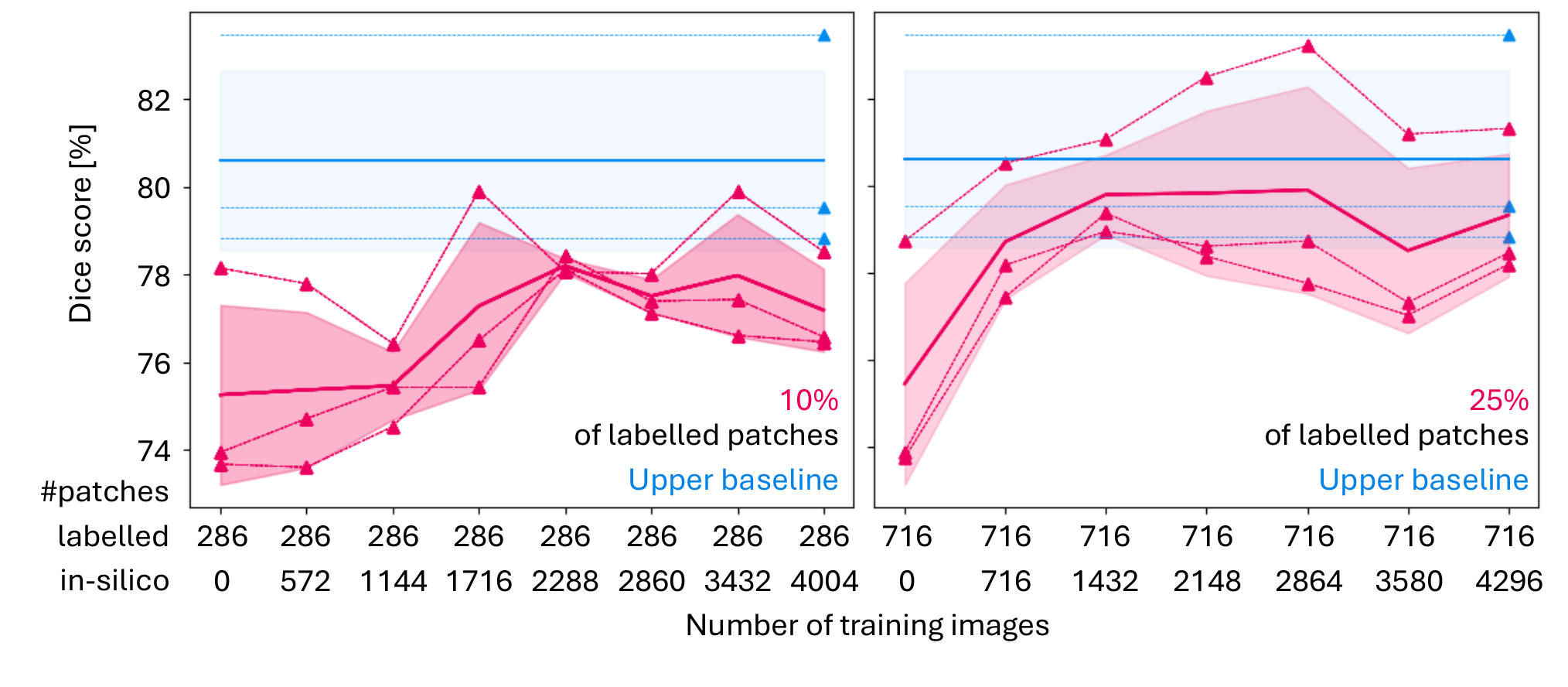}}
    \vspace*{-2mm}
    \caption{
      An ablation study was conducted to investigate performance saturation using 10\% ($N=286$ images) and 25\% ($N=716$ images) subsets. The Dice scores for the segmented epithelium regions, compared against annotations from three pathologists (denoted by triangles and dotted lines), are displayed alongside the corresponding mean (solid line) and standard deviation (shaded area).
    }
    \vspace*{-2mm}
  \label{figure3}
\end{minipage}
\end{figure}

\section{Experiments}
%\subsection{Data Generation}
The effectiveness of the proposed diffusion model for data augmentation was assessed in the context of epithelium region segmentation. Our training dataset comprised 61 Whole-Slide images (WSIs) of Non-Small Cell Lung Cancer (NSCLC) resections, stained with a nuclear marker assay (e.g. Ki67) and imaged with an Aperio AT2 scanner at 20x resolution. A total of 163 Field-of-Views (FOVs), each of size 2000x2000px, were positioned on these 61 WSIs, and epithelium regions were manually annotated by a pathologist. This resulted in 2865 patches (each 512x512px) in the training set and 1203 patches in a validation set, which were split on a WSI level. An independent test set comprising 43 FOVs, distributed over 18 WSIs of independent but similarly stained and scanned biopsy NSCLC samples, was annotated by three pathologists. 
We observed that the proposed diffusion model achieved optimal performance when the origin and target images shared similar area and shape characteristics of the foreground and background regions. Selecting masks for appearance transfer from the same dataset ensured similar class distributions in the target mask as in the source mask, as well as morphologically realistic structures in the target mask. The target masks were augmented with random flips and rotations.
\begin{figure}[htb]
  \begin{minipage}[b]{1.0\linewidth}
    \centering
    \centerline{\includegraphics[width=8.5cm]{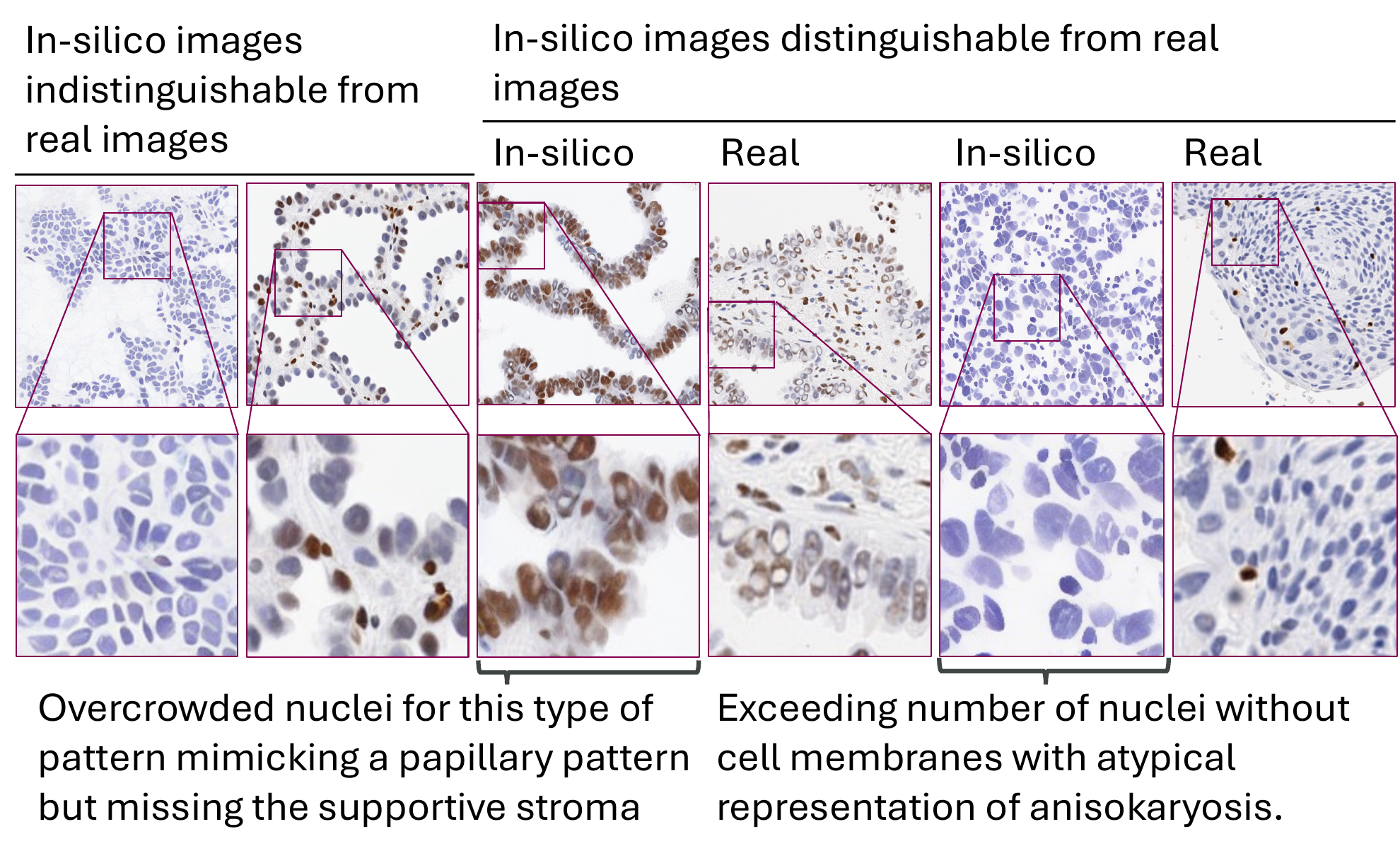}}
  \vspace*{-2mm}
  \caption{
  A board-certified pathologist assessed the differences between real and \textit{in-silico} images.
  }
  \vspace*{-2mm}
  \label{figure4}
\end{minipage}
\end{figure}
For the denoising process, we used the standard DDIM scheduler for 100 denoising steps \cite{song2020denoising}. Similar to the original work \cite{alaluf2023cross}, we injected the keys and values at timesteps between 10 and 70 for layers with a resolution of 32x32 and for timesteps between 10 and 90 for layers at resolution 64x64. Additionally, we applied the AdaIN operation at timesteps between 20 and 100 \cite{alaluf2023cross}. 
The out-of-the-box model from Alaluf et al. \cite{alaluf2023cross} is used as baseline transfer model. 
The downstream segmentation model consists of a U-Net with an Efficient Net backbone. During training we optimize the combined binary cross entropy and dice loss using the ADAM optimizer, with a learning rate of 0.001, $\beta_1$=0.9 and $\beta_2$=0.99. The best segmentation models were selected for each method independently as to maximize the dice score on the validation set. The selected segmentation models were evaluated on our held-out test set.
\section{Results}
In a series of quantitative experiments, we evaluated the utility of the generated in-silico data by applying it to a semantic epithelium segmentation task. The mean Dice scores, along with the associated low and high confidence intervals obtained through bootstrapping (999 resamples), are reported as follows:
\begin{enumerate}[label=(\alph*)]
  \item The upper baseline model, trained on all 2865 annotated patches, achieved a dice score of (80.59\%, [77.35\%, 83.13\%]; mean, [low, high confidence interval]). 
  \item The lower baseline model, trained on only 25\% (N=716) of the annotated patches yields a dice score of (75.42\%, [71.44\%, 78.64\%]).
  \item The model trained on a combination of 25\% (N=716) of the annotated patches and four \textit{in-silico} images generated for every annotated image with the proposed diffusion model yields a dice score of (79.88\%, [76.54\%, 82.51\%]).
  \item The model trained on a combination of 25\% (N=716) of the annotated patches and four \textit{in-silico} images generated for every annotated image with the baseline diffusion model yields a dice score of (75.09\%,[78.49\%, 81.21\%]).
\end{enumerate}
Wilcoxon signed rank tests \cite{conover1999practical} revealed a significant difference (p=0.038) between the lower baseline model (b) and the model with augmented data (c), but not between the latter and the upper baseline model (a) (p=0.23). Additionally, Wilcoxon signed rank tests show a significant difference (p=0.04) between the lower baseline model (b) and the baseline model [6] including augmented data (d) and between the latter (d) and the upper baseline model (a) (p=$2.8 \times 10^{-5}$).
Furthermore, we find that each fold of added data from our model yields significant improvements over the lower baseline segmentation model (p$<$0.05, Wilcoxon signed-rank tests, \ref{figure2}c).
The proposed diffusion model demonstrated an improvement over the baseline diffusion model. Performance achieved with the latter saturated after one-fold of \textit{in-silico} data generation, while further performance gains were observed with the proposed model (see Fig. \ref{figure2}b,c).
An ablation study, in which the dataset was reduced to 25\% and 10\% of its original size (Fig. \ref{figure3}), demonstrated the impact of the number of augmentation folds on segmentation performance. Accuracy saturated after four and eight folds of augmentation, respectively.
In a series of qualitative experiments, we assessed the accuracy of the generated IHC images with a board-certified pathologist. The goal was to identify potential areas for improvement. While the pathologist found that most morphologies were realistically generated, some discrepancies were noted (Fig. \ref{figure4}). These included overcrowded nuclear regions with unrealistic morphologies for NSCLC, resembling a papillary pattern without the corresponding stroma. Additionally, regions with extremely high nuclear density lacked cell membranes and exhibited atypical variations in nuclear size and shape (anisokaryosis).
 
\subsection{Discussion}
In this study, we introduce a novel diffusion model-based data generation pipeline optimized for computational pathology. The model, originally trained on natural images, is adapted and applied in a zero-shot setting. By extending a baseline diffusion model to a multi-class framework, we can generate realistic IHC images, reducing reliance on labeled data by approximately 75\% without compromising performance in downstream segmentation tasks. 
In collaboration with a certified pathologist, we found that real and \textit{in-silico} images are primarily distinguishable based on nuclear morphology. This observation suggests the potential for training a histopathology-specific backbone to improve model performance. However, we acknowledge that our approach requires segmentation masks and is not fully applicable in a fully unsupervised setup without annotations. 
Further investigation is required to assess the model’s performance in scenarios involving more than two classes or inaccurate annotations. Additionally, we plan to compare our approach with the Semantic Diffusion Model \cite{wang2022semantic} to evaluate its relative strengths. We also aim to explore how \textit{in-silico} data can help mitigate performance gaps in deep learning models for computational pathology, especially in cases of imbalanced datasets \cite{vaidya2024demographic}. We believe this study highlights the effective use of zero-shot models for generating high-quality \textit{in-silico} data in computational pathology.

\section{Compliance with ethical standards}
\label{sec:ethics}
The study was conducted in adherence with the International Council for Harmonization Good Clinical Practice guidelines, the Declaration of Helsinki, and local regulations on the con- duct of clinical research.

\section{Acknowledgments}
\label{sec:acknowledgments}
All authors are employees of AstraZeneca and do not have other relevant financial or non-financial interests to disclose.
% Below is an example of how to insert images. Delete the ``\vspace'' line,
% uncomment the preceding line ``\centerline...'' and replace ``imageX.ps''
% with a suitable PostScript file name.
% -------------------------------------------------------------------------


\bibliographystyle{IEEEbib}
\bibliography{bibliography}
\end{document}